%% file: main.tex
\documentclass[letterpaper, 10 pt, conference]{style/ieeeconf}

\IEEEoverridecommandlockouts
\usepackage{graphics} 
\usepackage{epsfig} 
\usepackage{mathptmx} 
\usepackage{times} 
\usepackage{amsmath} 
\usepackage{amssymb}  
\usepackage{siunitx} 
\usepackage[inkscapelatex=false]{svg}

\usepackage[hidelinks]{hyperref}
\usepackage{subcaption} 
\usepackage{pdfpages} 

\usepackage{algorithm}
\usepackage{algpseudocode}

\usepackage{authblk}

\usepackage{calc}
\usepackage{array}
\usepackage{tabulary}
\usepackage{booktabs}
\usepackage{multirow}

\DeclareMathAlphabet{\mathcal}{OMS}{cmsy}{m}{n}

\usepackage{mathtools}

\title{\LARGE \bf
Learning Perceptive Platform Adaptive Locomotion Controllers for Quadrupedal Robots}

\author[1]{David Rytz\thanks{
Corresponding author: rytz@robots.ox.ac.uk. Video can be found \color{blue}{\href{https://drive.google.com/file/d/19q2Js218BH0w7772GMsqNwLCGlGbBZTo/view?usp=sharing}{here}}.
}
}
\author[1]{Kim Tien Ly}
\author[1]{Ioannis Havoutis}
\affil[1]{Dynamic Robot Systems, Oxford Robotics Institute, University of Oxford}
\begin{document}
%
\maketitle

\input{sections/abstract}


\input{sections/introduction}

\input{sections/methodology}

\input{sections/results_and_discussion}


\bibliographystyle{style/IEEEtran}

\bibliography{references}




\end{document}

%% file: sections/abstract.tex
\begin{abstract}%
Universal quadrupedal locomotion remains limited by the difficulty of integrating perception across diverse robot morphologies. State-of-the-art controllers rely on single-robot training or blind policies that omit real-time perception, leading to poor cross-embodiment generalization. Designing locomotion policies that remain robust across related quadruped morphologies while incorporating perception is challenging. Moreover, fully perceptive policies are often sensitive to noise, whereas blind controllers lack terrain awareness. In this work, we study how perception should be integrated into morphology-aware reinforcement learning architectures for deployable quadrupedal control. Building on MorAL, we train morphology-specialized universal controllers on multiple reference quadrupeds using adaptive terrain curricula. We compare a blind baseline, a critic-perceptive variant (MorAL+), and a fully perceptive actor–critic (PPAL). Policies are evaluated in simulation on flat and rough terrains, and deployed on ANYmal hardware. Results show that critic-only perception improves robustness and tracking consistency over blind baselines while remaining more stable than fully perceptive policies under perception noise. These findings highlight that perception placement and curriculum design are key factors for scalable, morphology-aware locomotion.

\end{abstract}

%% file: sections/introduction.tex
\section{Introduction}

The ability to curate and process large datasets has driven a significant push for general-purpose artificial intelligence (AI) models that generalize across diverse information modalities, including images, video, audio, and more. This development has led to recent cross-embodiment models, such as Vision-Language-Action (VLA) generalist models, exemplified by Gemini Robotics~\cite{teamGeminiRoboticsBringing2025}, which can be fine-tuned to control various real robot modalities. To enable similar large-scale mobile robot control, LocoFormer~\cite{liuLocoFormerGeneralistLocomotion2025} demonstrated the ability to synthesize \textit{universal} omni-bodied locomotion models, able to control a wide range of robot bodies with significant changes, including locked or damaged limbs. The key to such generalist capability is massive domain randomization using procedurally generated robots with diverse robot embodiments across bipeds, quadrupeds, and wheeled varieties, while also randomizing relevant parameters, including joint control gains, body masses, and inertias~\cite{liuLocoFormerGeneralistLocomotion2025}. This approach stands in contrast to the very capable Reinforcement Learning (RL) methods for training locomotion controllers~\cite{zhuangRobotParkourLearning2023, hoellerANYmalParkourLearning2024, kimHighspeedControlNavigation2025}, which are trained specifically for a single robot and thus reduce their usefulness across different robot parameters and manufacturers, limiting scalability and knowledge reuse. 

In this work, we pursue a middle ground: morphology-specialized universal locomotion controllers that incorporate perception while maintaining tractable computational demands. Building on MorAL~\cite{luoMorALLearningMorphologically2024}, we reformulate universal locomotion as a morphology-specific learning problem with the goal of perceptive locomotion without requiring large-scale cross-embodiment training~\cite{liuLocoFormerGeneralistLocomotion2025, bohlingerOnePolicyRun2025, yangMultiLocoUnifyingMultiEmbodiment2025}. This formulation allows efficient deployment and transfer to real hardware while preserving terrain awareness and robustness. Our key contributions are:
\begin{itemize}
    \item We conduct a study of morphology-aware blind, critic-perceptive, and fully perceptive actor-critic locomotion architectures to develop a perceptive and single-morphology \textit{universal} locomotion policy for quadrupedal robots with diverse kinematic and dynamic properties.
    \item We design an adaptive terrain curriculum that balances sample efficiency with challenging environmental variability, improving robustness and generalization.
    \item We extend the MorAL architecture to support multi-reference morphology training with perception-aware critic information, enabling efficient zero-shot transfer to unseen quadruped systems in simulation and hardware.
\end{itemize}

\section{Related Work}\label{s:related_work}

\subsection{Reinforcement Learning for Quadrupedal Locomotion}
Reinforcement Learning (RL) has shown in recent years to be a powerful tool to generate robust and agile legged locomotion controllers by training policies in simulation with extensive domain randomization and subsequently transferring them zero-shot to real hardware. Such strategies enable policies to adapt to unmodeled dynamics and environmental variations through implicit experience during training, reducing the need for fine-tuning after deployment~\cite{leeLearningQuadrupedalLocomotion2020, zhangLearningAgileLocomotion2024}.

Incorporating perception into the control loop enables terrain-aware adaptation. These perceptive locomotion approaches allow robots to anticipate and adjust their gait based on the upcoming terrain geometry. Depending on how visual information is represented, methods can be categorized into either \textit{implicit} or \textit{explicit} perception frameworks. Implicit methods~\cite{agarwalLeggedLocomotionChallenging2022, chengExtremeParkourLegged2024} integrate visual or egocentric features directly into the policy network, enabling end-to-end learning of terrain-adaptive behaviors. In contrast, explicit mapping approaches~\cite{kimHighspeedControlNavigation2025, mikiLearningRobustPerceptive2022} rely on geometric or elevation-based representations to inform footstep planning and control.

\subsection{Universal and Cross-Embodiment Locomotion Control}

Cross-embodiment learning seeks to generalize locomotion control across robots with distinct morphologies, actuation schemes, and sensor configurations. Early research emphasized morphology-specialized universal controllers—policies trained for robustness across a family of related morphologies rather than full embodiment agnosticism.

GenLoco~\cite{fengGenLocoGeneralizedLocomotion2022} trained quadruped controllers through procedural morphology randomization and a large context window, achieving robust behavior across varied parameters but limited to fixed degrees of freedom (DoFs). ManyQuadrupeds~\cite{shafieeManyQuadrupedsLearningSingle2023} combined Central Pattern Generators (CPGs) with RL for multi-quadruped control but relied on robot-specific inverse kinematics, restricting scalability. MorAL~\cite{luoMorALLearningMorphologically2024} advanced this paradigm via a morphology-aware architecture encoding physical information from proprioceptive inputs into compact morphological representations, enhancing robustness across diverse quadruped configurations. All of these approaches demonstrate rough terrain locomotion capabilities on small quadrupeds below~\SI{30}{\kilo\gram}.

Broader cross-embodiment approaches such as URMA~\cite{bohlingerOnePolicyRun2025} and LocoFormer~\cite{liuLocoFormerGeneralistLocomotion2025} pursue morphology-agnostic architectures using large-scale transformer encoders and procedurally generated robot datasets. URMA unifies control across quadrupeds, bipeds, and hexapods via joint-level descriptors, while LocoFormer trains an omnibodied transformer model across dozens of GPUs with extended temporal context. Although these methods achieve strong embodiment generalization, their massive computational demands limit practicality for real-world deployment.

Building upon MorAL, our work introduces morphology-specialized universal architectures that retain perceptive locomotion capabilities with substantially reduced computational requirements compared to cross-embodiment models. MorAL utilizes a single robot reference model augmented with heightmap-based perceptual encoding. We extend the asymmetric actor–critic design, where only the critic receives exteroceptive input during training, to a multi-reference setup. This formulation unifies morphology encoding and minimal perception injection, bridging morphology-specific generalization and terrain-aware control for scalable, zero-shot deployment across diverse and previously unseen quadrupedal systems while maintaining deployment simplicity.

%% file: sections/methodology.tex
\section{Methodology}
\label{s:methodology}

Our goal in this work is to train a \textit{universal} perceptive locomotion policy that can robustly control a range of previously unseen quadrupedal robots in a zero-shot fashion to follow a given base velocity command, consisting of forward and lateral velocities and yaw rate. The command is uniformly sampled within the training ranges $v_x^\text{max} = \pm \SI{1}{m/s}$, $v_y^\text{max} = \pm \SI{0.75}{m/s}$, and $\omega_z^\text{max} = \pm \SI{1.5}{rad/s}$.

\subsection{Problem Formulation}
We model the problem as a partially observable Markov decision process (POMDP), represented by a 7-tuple 
$M = \{S, O, A, R, T, \Omega, \gamma\}$. Here, $S$, $O$, and $A$ denote the sets of states, observations, and actions, respectively. At each time step $t$, the agent in state $s_t \in S$ selects an action $a_t \in A$ and receives a reward $R(s_t, a_t)$. 
The environment then transitions to a new state $s_{t+1}$ with probability $T(s_{t+1} \mid s_t, a_t)$. Because the full state is not directly observable (e.g., complete terrain information), the agent instead receives an observation $o_{t+1} \in O$ generated according to the observation function $\Omega(o_{t+1} \mid s_{t+1}, a_t)$. The scalar $\gamma$ denotes the discount factor. 

The objective is to find the optimal policy $\pi^*$ that maximizes the expected discounted cumulative reward:
\begin{equation}
    J_M(\pi) = \mathbb{E}_{\pi} \left[ \sum_{t=0}^{\infty} \gamma^t R(s_t, a_t) \right].
\end{equation}
While previous works often focused more on robot-specific terrain information processing~\cite{mikiLearningRobustPerceptive2022} or massive cross-embodiement configurations~\cite{liuLocoFormerGeneralistLocomotion2025, bohlingerOnePolicyRun2025}, we aim to also deploy on heavy hardware, such as the \SI{50}{\kilo\gram} ANYmal robot, and thus investigate asynchronous actor-critic network layouts as described in Section~\ref{ss:observation_state_space}, inspired by~\cite{luoMorALLearningMorphologically2024,rytzSamplingStrategiesRobust2025}.

\subsection{Controller and Training Design}\label{ss:training}
\begin{figure*}[h!]
    \centering
    \includegraphics[width=0.8\linewidth]{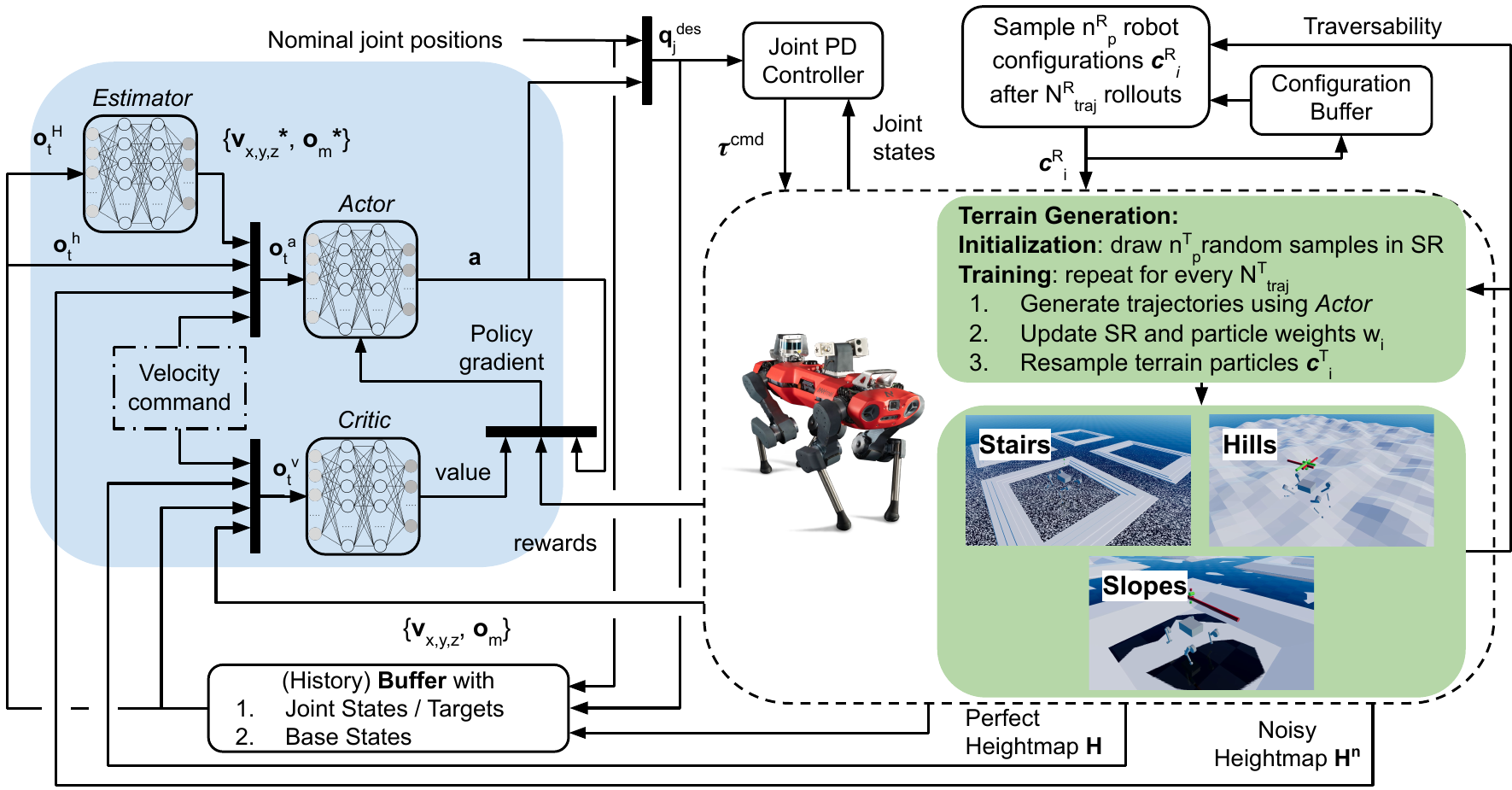}
    \caption{Architecture for perceptive \textit{universal} quadrupedal locomotion: The base and joint states and actions are stored in a buffer. The estimator network receives $\mathbf{o}_t^H$, a history of robot states from the buffer. The critic receives the observation $\mathbf{o}_t^v$ including the perfect heightmap $\textbf{H}$ as input. The Actor receives $\mathbf{o}_t^a$, including the corresponding estimator outputs, and the noisy heightmap $\textbf{H}^n$ and is finally deployed on the ANYmal platform. While the robot parameter configuration follows~\cite{rytzReferenceFreePlatform2025}, we include work terrain generation (in green) as outlined in Section~\ref{ss:configParamSampling}.}
    \label{fig:methodology:control_framework}
    \vspace{-0.6cm}
\end{figure*}

An overview of our proposed control framework is illustrated in Fig.~\ref{fig:methodology:control_framework}. The framework consists of three different neural networks: an estimator, a critic, and an actor. The estimator network estimates relevant state and robot morphology parameters for control using proprioceptive quantities and feeds them to the actor network, which outputs joint target commands that are then tracked using a joint proportional-derivative (PD) controller. The critic network helps reduce variance in the policy gradient estimate from RL algorithms. All training is done in simulation, using RaiSim~\cite{raisimtechArticulatedSystemsRaiSim}.

\subsubsection{Observation and State Space}\label{ss:observation_state_space}
To train robot locomotion policies for rough-terrain traversal, our framework uses a heterogeneous actor-critic architecture. The actor network generates control policies that output joint target commands, and the critic evaluates the quality of these policies while receiving privileged information. Together with the morphology and base estimator network, we can adapt to a wide range of robots and terrain during deployment. These three networks are trained in parallel in simulation, with the estimator being updated via a regression algorithm. We discuss each network in more detail here:

\begin{enumerate}
    \item \textbf{Estimator Network:} 
        In the absence of imitation learning, the framework requires the estimator to infer the base linear velocity $\mathbf{v}_\text{x,y,z}^*$ to enable base velocity command tracking. To enhance generalization across different morphologies, this estimator network is augmented with robot-specific morphological descriptors, $\mathbf{o}_m^*$. These descriptors include the base body center of mass $\textbf{com}_{\text{base},{x,y,z}}\in\mathbb{R}^3$, the individual link masses $\mathbf{m}_\text{base, hip, thigh, shank}\in\mathbb{R}^4$, joint link offsets $\mathbf{c}_{q_1^\text{xyz}, q_2^\text{xyz}, q_3^\text{xyz}}\in\mathbb{R}^9$, and the vertical foot offset $c_{f,z}\in\mathbb{R}^1$.  The estimator operates on a short temporal context defined by the robot state history $\mathbf{o}_t^H = \langle \mathbf{o}^h_{t=0}, \mathbf{o}^h_{t=1}, \ldots, \mathbf{o}^h_{t=5} \rangle$ with each historical state vector encapsulating proprioceptive and command information. For the current timestep $t=0$, the observation vector is specified as  
        \begin{equation}
            \mathbf{o}_{t=0}^h = \langle \mathbf{e}_z^B, \mathbf{\omega}_B, \Delta\mathbf{q_t}, \mathbf{\dot{q}_t}, \mathbf{q_{t-1}^\text{des}}, \mathbf{v}_{x,y,\omega}^\text{des} \rangle,
        \end{equation}
        where $\mathbf{e}_z^B\in\mathbb{R}^3$ denotes the base orientation unit vector, $\mathbf{\omega}_B\in\mathbb{R}^3$ represents the base angular velocity, $\Delta\mathbf{q_t}\in\mathbb{R}^{12}$ corresponds to the deviation of joint positions from their nominal configuration, $\mathbf{\dot{q}_t}\in\mathbb{R}^{12}$ is the joint velocity vector, $\mathbf{q_{t-1}^\text{des}}\in\mathbb{R}^{12}$ is the previously executed joint action, and $\mathbf{v}_{x,y,\omega}^\text{des}\in\mathbb{R}^3$ specifies the commanded velocity in heading, lateral and yaw directions. For our study, the history is selected at multiples of \SI{0.01}{\second}.
        
    \item \textbf{Actor Network:}
        The actor network produces the desired joint position command $\mathbf{q}_j^\text{des}\in\mathbb{R}^{12}$ by processing an input observation vector $\mathbf{o}^a$ and outputting the desired action $\mathbf{a}_t\in\mathbb{R}^{12}$. This input encapsulates the estimated base linear velocity $\mathbf{v}_\text{x,y,z}^*$ and morphology descriptor $\mathbf{o}_m^*$, 
        the current robot state $\mathbf{o}_{t=0}^h$, the base centered noisy height scan $\mathbf{H}^n\in\mathbb{R}^{108}$, and the nominal joint configuration $\mathbf{q}^n\in\mathbb{R}^{12}$, defined as
        \begin{equation}
            \mathbf{o}_t^a = \langle \mathbf{v}_\text{x,y,z}^*, \mathbf{o}_{t=0}^h, \mathbf{q}^\text{n}, \mathbf{H}^n, \mathbf{o}_m^* \rangle.
        \end{equation}
        The resulting action is drawn from a Gaussian distribution with zero mean and a fixed standard deviation. The sampled output $\mathbf{a}$ is added to the nominal joint position $\mathbf{q}^\text{n}$ before being applied to the robot’s joint-space PD controller, expressed as $\mathbf{q}^{des} = \mathbf{q}^\text{n} + \sigma_a \cdot \mathbf{a}$ with $\sigma_a = 0.6$.
        
    \item \textbf{Critic Network:}
        The critic network is given access to additional privileged information, drawn from the simulator that is generally unavailable or difficult to estimate during real-world deployment. This privileged data includes the feet contact states $\mathbf{c}_t^{f, i}\in\mathbb{R}^4$, the corresponding contact friction coefficients $\mathbf{\mu}^{f, i}\in\mathbb{R}^4$, the individual feet height in base frame $\mathbf{h}_{t}^{f, i}$, the joint-space PD gains $\mathbf{k}_{PD}\in\mathbb{R}^{24}$, and the noise-free height scan $\mathbf{H}\in\mathbb{R}^{108}$. Unlike the actor, which uses estimated morphological parameters, the critic processes ground-truth morphological descriptors $\mathbf{o}_m$ obtained directly from the simulation environment. The resulting input observation for the critic is thus expressed as      
        \begin{equation}
            \mathbf{o}_t^v = \langle \mathbf{v}_\text{x,y,z}, \mathbf{o}_{t=0}^h, \mathbf{q}^n, \mathbf{o}_m, \mathbf{c}_t^{f, i}, \mathbf{\mu}^{f, i}, \mathbf{h}_{t}^{f, i}, \mathbf{k}_{PD}, \mathbf{H} \rangle.
        \end{equation}

\end{enumerate}


The estimator network is optimized by a supervised regression loss $loss_{reg}$, similar to~\cite{luoMorALLearningMorphologically2024}, to reduce the mean square error (MSE) between the estimation and corresponding ground-truth values. The total loss is computed as:
\begin{equation}
    Loss = \beta\left( MSE(\mathbf{v}_{x,y,z},\mathbf{v}^*_{x,y,z}) + MSE(\mathbf{o}_m, \mathbf{o}_m^*) \right) + (1-\beta) loss_\pi,
\end{equation}
with the hyperparameter $\beta = 0.04$ and $loss_\pi$ being the policy gradient. 

\begin{figure}
    \centering
    \includegraphics[width=0.63\linewidth]{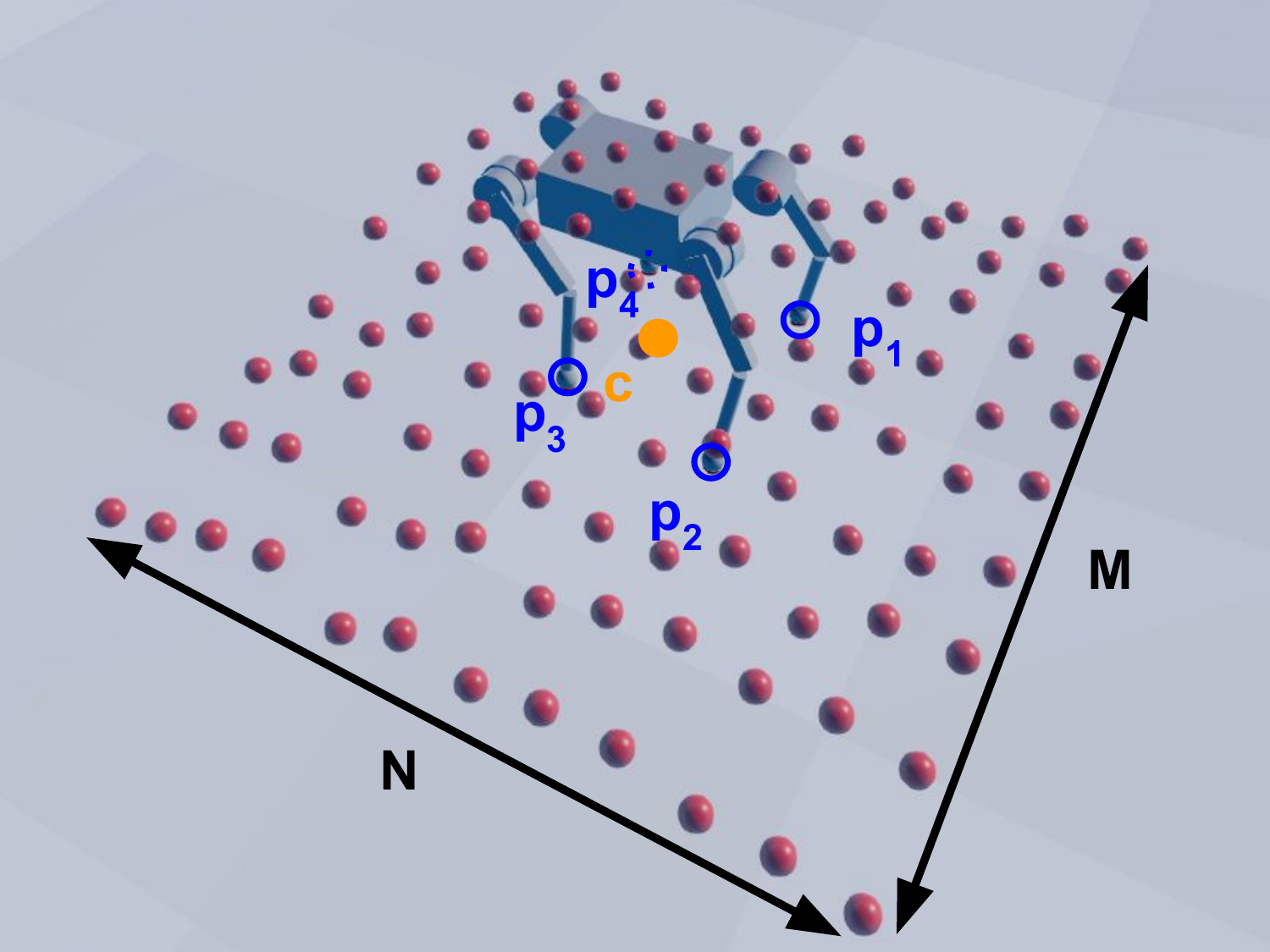}
   \caption{Robot-centric heightmap scan pattern $\mathbf{H}^n$, displayed as N$\times$M red dots. During training, an additional noise value is added sampled as $h^n \sim \mathcal{U}(0.02, 0.1)$.} 
    \label{fig:methodology:height_map_scan}
    \vspace{-0.5cm}
\end{figure}

We dynamically generate a robot-centric grid of $12\times9$ sampling points centered around the estimated support polygon of the robot’s feet in the nominal robot joint configuration (see Fig.~\ref{fig:methodology:height_map_scan}) inspired by~\cite{rudin2022}. It shows in red an example noisy heightmap scan pattern for an A1 reference model. We include latency randomization during training for the perception observations $\mathbf{H}$ and $\mathbf{H}^n$. 

\subsubsection{Rewards}
We detail the reward components and their corresponding coefficients in Table~\ref{tab:fp_reward_definitions}, following the formulations in~\cite{luoMorALLearningMorphologically2024, rytzReferenceFreePlatform2025}. Unless otherwise specified, all policies in this study were trained using the same reward structure and weights.

\begin{table}[h]
    \centering
    \footnotesize  
    \setlength{\tabcolsep}{4pt}  
    \renewcommand{\arraystretch}{1.0}  
    \begin{tabular}{|l|l|}
        \hline
        \textbf{Reward} & \textbf{Description} \\
        \hline
        Base linear velocity & $\mathrm{r}_{v} = 3 \cdot (1-\tanh(4\Vert \mathbf{c}_{xy}^\text{des} - \mathbf{c}_{xy} \Vert^2))$ \\
        \hline
        Base angular velocity & $\mathrm{r}_\omega = 1.5 \cdot (1-\tanh(2\Vert \mathbf{c}_{z}^\text{des} - \mathbf{c}_{z} \Vert^2))$ \\
        \hline
        Base orientation & $\mathrm{r}_R = -5 \cdot \tanh(R_{B})^2$ \\
        \hline
        Base height & $\mathrm{r}_h = -20 \cdot \tanh((r_z - r_n)^2)$ \\
        \hline
        Base undesired motion & $\mathrm{r}_b = -0.5 \cdot (v_z^2 + 0.25 \cdot (|\omega_x| + |\omega_y|))$ \\
        \hline  
        Joint position & $\mathrm{r}_q = -0.2 \cdot ||q_t - q^\text{n}||^2$ \\
        \hline
        Joint velocity & $\mathrm{r}_{\dot{q}} = -3 \cdot 10^{-4} \cdot \Vert \dot{q}_t \Vert^2$ \\
        \hline
        Joint acceleration & $\mathrm{r}_{\ddot{q}} = -1 \cdot 10^{-7} \cdot \Vert \ddot{q}_t \Vert^2$ \\
        \hline
        Joint torque & $\mathrm{r}_{\tau} = -3.5 \cdot 10^{-5} \cdot \Vert \tau \Vert^2$ \\
        \hline
        Joint action smoothness & $\mathrm{r}_{s1} = -0.12 \cdot ||q_t^{des} - q_{t-1}^{des}||^2$ \\
        \hline
        Joint action smoothness 2 & $\mathrm{r}_{s2} = -0.05 \cdot ||q_t^{des} - 2\cdot q_{t-1}^{des} + q_{t-2}^{des}||^2$ \\
        \hline
        Foot slip & $\mathrm{r}_{\mu,i} = -0.25 \cdot c_{f,i} \cdot \Vert v_{f,xy,i} \Vert$ \\
        \hline
        Air time & $\mathrm{r}_{air,i} = -6 \cdot \begin{cases}
            \text{if}~||\mathbf{c}^\text{des}||^2 = 0: & -T_{i,swing} \\
            \text{else:} & T_{i,swing} - 0.5
        \end{cases}$ \\
        \hline 
    \end{tabular}
    \caption{Reward term definitions.}
    \label{tab:fp_reward_definitions}
    \vspace{-0.3mm}
\end{table}

The index $i\in\left\{1,2,3,4\right\}$ represents the corresponding leg (front left, front right, hind left, and hind right). $T_{i,stance}$ and $T_{i,swing}$ are the corresponding times since the last touchdown and takeoff. The contact state $\mathrm{c}_{f,i}$ is taken from the simulator for each foot $i$. If $\mathrm{c}_{f,i}=1$, we define foot $i$ to be in ground contact and 0 otherwise. $R_{B}$ represents the base orientation with respect to the world frame. The tangential foot velocity at its contact point is captured by $v_{f,xy,i}$. 

We would like to point out that \textit{Luo et al.}~\cite{luoMorALLearningMorphologically2024} successfully trained and deployed a perceptive locomotion policy for light and medium-sized robots of less than \SI{30}{\kilo\gram} without any foot clearance reward, which we aim to reproduce in this work.

\subsubsection{Joint controller}
As illustrated in Fig.~\ref{fig:methodology:control_framework}, each joint $j \in \{1, \ldots, 12\}$ is regulated using a PD controller that tracks the desired joint position $\mathrm{q}_j^{\text{des}}$ generated by the actor output $a_j$. The commanded torque $\tau_j^{\text{cmd}}$ applied to the actuator is defined as
\begin{equation}
    \tau_j^{\text{cmd}} = K_p \cdot (\mathrm{q}^{\text{des}}_j - \mathrm{q}_j)
    - K_d \cdot \mathrm{\dot{q}}_j,
    \label{eq:impedance_controller_simplified}
\end{equation}
where $K_p$ and $K_d$ denote the proportional and derivative gains, respectively. The commanded torques $\tau_j$ are clipped within the actuator limits $[-\tau^{\text{max}}_j, \tau^{\text{max}}_j]$. To capture more realistic actuation dynamics, which are typically nonlinear and complex in physical systems such as the series elastic actuators of the ANYbotics robots, we incorporate learned actuator models \cite{hwangboLearningAgileDynamic2019a}. The actuator network stochastically replaces the PD formulation in Equation~(\ref{eq:impedance_controller_simplified}) and is trained on data collected from the ANYmal~B and ANYmal~C hardware platforms.

\subsection{Terrain Configuration Parameters and Heightmap Sampling}\label{sec:appendix:terrainConfigParameters}

We generate an $N \times M$ grid to be sampled for heightmap values during training and on the off-the-shelf perception pipeline by ANYbotics as follows:
            
Given the set of foot nominal positions $\mathcal{F} = \{ \mathbf{p}_i \in \mathbb{R}^2 \}_{i=1}^{m=4}$, we compute the axis-aligned bounding box and its centroid $c$:
\begin{align*}
    \mathbf{p}_{\min} &= \min_i \mathbf{p}_i, &
    \mathbf{p}_{\max} &= \max_i \mathbf{p}_i, &
    \mathbf{c} &= \frac{1}{m}\sum_{i=1}^{m} \mathbf{p}_i.
    \end{align*}
To ensure sufficient coverage, the bounding box is expanded by user-defined paddings $p_x=$\SI{0.5}{\meter} and $p_y=$\SI{0.5}{\meter}:
\begin{equation}
    \mathbf{p}_{\min} \leftarrow \mathbf{p}_{\min} - [p_x, p_y]^T, \quad
    \mathbf{p}_{\max} \leftarrow \mathbf{p}_{\max} + [p_x, p_y]^T.            
\end{equation}
The grid spacing along each axis is determined as
\begin{equation}
    \Delta x = \max\big(\mathrm{round}_\text{\SI{1}{\centi\meter}}\!\big(\tfrac{p_{\max,x} - p_{\min,x}}{N-1}\big),\ s_\text{min}\big),
    \quad
\end{equation}
where $\mathrm{round}_\text{\SI{1}{\centi\meter}}(\cdot)$ enforces quantization to 1\,cm resolution and $s_\text{min}$ denotes the minimum allowable spacing, set to \SI{2}{\centi\meter}. The final grid is centered at $\mathbf{c}$:
\begin{equation}
    x_i = c_x - \tfrac{(N-1)\Delta x}{2} + i\,\Delta x, \quad
    y_j = c_y - \tfrac{(M-1)\Delta y}{2} + j\,\Delta y,            
\end{equation}
for $i \in [0, N-1]$ and $j \in [0, M-1]$, producing a total of $N\times M$ sampling points
\begin{equation}
    \mathbf{s}_{(i,j)} = (x_i, y_j).    
\end{equation}


\subsection{Parameter Sampling Strategies for Universal Quadrupedal Robot Control}\label{ss:configParamSampling}

We employ two strategies for sampling $n_p^R$ robot configuration parameters $\mathbf{c}^R_i$ and $n_p^T$ terrain configurations parameters $\mathbf{c}^T_i$ during training after $N_\text{traj}^R$ or $N_\text{traj}^T$ seconds of simulator rollouts:

\begin{itemize}
    \item \textbf{Adaptive curriculum with particle filtering:} 
        Inspired by~\cite{leeLearningQuadrupedalLocomotion2020}, we define the linear and zero-velocity tracking performance indicators as:
        %
            \begin{align}
        \nu^{\text{lin}}(t) &= 
        \begin{cases} 
            1 & \text{if } |\nu_{pr}^{\text{lin}}| > 0.3 \cdot \| [v_x^{\text{max}}, v_y^{\text{max}}] \| \\ 
            0 & \text{otherwise} 
        \end{cases} \\
        \nu^{\text{zero}}(t) &= 
        \begin{cases} 
            1 & \text{if } \|\mathbf{v}_{x,y,\omega}^{\text{cur}}\| < 0.2 \\ 
            0 & \text{otherwise} 
        \end{cases}
            \label{eq:tracking}
    \end{align}
        Here, $\nu_{pr}^{lin}$ is the scalar projection of the commanded onto the current base linear velocity in the robot frame, and $|\mathbf{v}_{x,y,\omega}^\text{cur}|$ denotes the velocity norm during zero-velocity commands. 
        Over all timesteps, the tracking performance for configuration $i$ is computed as
        \begin{equation}
            Tr_i^\text{cmd} = \sum_{t=0}^{T_\text{tot}} \frac{\nu_i^{cmd}(t)}{T_i^\text{tot}},
        \end{equation}
        where $T_\text{tot}$ is the total duration that a command is active for configuration $i$ after $N_\text{traj}$. Ideal tracking yields $Tr_i^\text{cmd}=1$ for $\text{cmd}\in\{\text{lin, zero}\}$, while poor policies approach zero. A Sequential Importance Resampling (SIR) particle filter maintains a parameter distribution that evolves with policy performance. Each configuration $i$ is assigned an importance weight
        \begin{equation}
            w_i = \frac{1}{2} \sum^{cmd} Tr_i^\text{lin}\in[0.45,0.95] + \gamma \cdot Tr_i^\text{zero}\in[0.45,0.95],
            \label{eq:w_k_i_particleUp}
        \end{equation}
        emphasizing configurations with mid- to upper-range performance. Weighted resampling, combined with random nearest-neighbor walks over $n_w$ configurations, prevents parameter set degeneration. $\gamma$ is a hyperparameter set to 1 for robot configuration sampling and 0 for terrain sampling.
    \item \textbf{Performance-based curriculum:} As per \cite{rytzSamplingStrategiesRobust2025}, 
        $n_\text{SR}$ configurations are generated from nominal parameters $c_i^\text{nom}$ by restricting the uniform sampling range (SR) to $SR\in(0.1,1)$, initialized at $0.1$. When the policy surpasses task-specific thresholds after each $N_\text{traj}$, $SR$ is adjusted to match training difficulty, based on the average tracking performance of all robot particles belonging to the same reference model. Let $Tr_\text{mean}^\text{cmd}$ denote the mean tracking score computed over all configurations $\mathbf{c}_i$. The update rule is defined as:
        \begin{equation}
            SR_{t+1} =
            \begin{cases}
                SR_{t} + 0.01, & \text{if } Tr_{\text{mean}}^{\text{lin}} > Tr_{\text{high}}^{\text{lin}}, \\[6pt]
                SR_{t} - 0.01, & \text{if } Tr_{\text{mean}}^{\text{lin}} < Tr_{\text{low}}^{\text{lin}}, \\[6pt]
                SR_{t}, & \text{otherwise.}
            \end{cases}
            \label{eq:sr}
        \end{equation}
        This adaptive scheme progressively increases task difficulty in proportion to learning progress, avoiding degenerate, catastrophic forgetting or overly trivial parameterizations $\mathbf{c}_i$.
\end{itemize}


\subsubsection{Robot Parameters}
We implement the robot parameter sampling approach by Rytz et al.~\cite{rytzSamplingStrategiesRobust2025}, using four simplified quadruped reference models that are employed during training: Unitree’s A1~\cite{unitreeA12023}, Aliengo, and ANYbotics’ ANYmal~B and ANYmal~C~\cite{hutterANYmalHighlyMobile2016a, ackermanevanANYboticsIntroducesSleek2019}. Nominal parameters from these platforms are used to resample new parameter sets $\mathbf{c}_i^R$, generating $n_p^R=30$ distinct configurations per model through randomized kinematic and dynamic properties. The sampling parameter ranges are set as in the original work of \textit{Rytz et al.}. Based on the \textit{reference quadrupeds}, $n_w = 0.1$  and $n_\text{SR}=0.4$ morphology parameters are sampled after $N_\text{traj}^R=$\SI{20}{\second}. The $SR$ is fixed to 1 at all times. A sampled robot configuration is admitted into the randomized training set if it can stand collision-free in its nominal pose for \SI{2}{\second} using RaiSim’s~\cite{raisimtechArticulatedSystemsRaiSim} joint controller. 


\subsubsection{Terrain Parameters}
To enable perceptive heavy-duty \textit{universal} morphology-adaptive locomotion~\cite{rytzSamplingStrategiesRobust2025}, we extend the framework of~\cite{luoMorALLearningMorphologically2024}: instead of a single curriculum per reference robot, each configuration follows an adaptive terrain curriculum, reflecting that robots with different morphologies possess distinct terrain capabilities (e.g., larger robots with higher torque limits might handle steeper terrain). Our terrain generation incorporates slope, stair, and hilly terrain, previously shown to be effective for rough-terrain control~\cite{leeLearningQuadrupedalLocomotion2020, rudin2022}.
    
Terrain parameters, defined in Table~\ref{tab:terrain_parameters}, are uniformly sampled within preset bounds with $SR=0.1$ at the start of each episode to enhance policy generalization. Depending on their respective traversability defined in Equation~(\ref{eq:tracking}), we increase or decrease $SR$ following Equation~(\ref{eq:sr}), with $Tr_\text{high}^\text{lin}=0.7$, and $Tr_\text{low}^\text{lin}=0.55$. We generate $n_w = 0.4$  and $n_\text{SR}=0.1$ samples after $N_\text{traj}^T=$\SI{20}{\second} for the total of 12 configurations per terrain and robot type. A flat terrain is sampled with a probability of \SI{10}{\percent}, all others with \SI{30}{\percent}. The resulting tiles for stair and slope terrains are replicated in a $3\times3$ grid with small flat gaps between tiles to create extended patterns. All generated heightmaps are centered such that the robot spawns on a flat central surface. 
\begin{table}[h]
\centering
\setlength{\tabcolsep}{6pt} 
\begin{tabular}{|l|l|c|c|}
    \hline
    \textbf{Terrain} & \textbf{Parameter} & \textbf{Sampling Distribution} & \textbf{Nominal} \\
    \hline
    \multirow{4}{*}{\textbf{Slope}} 
        & Step Size           & $\mathcal{U}(0.10, 1.00)$   & 0.40 \\ 
        & Max Height          & $\mathcal{U}(0.01, 0.30)$   & 0.01 \\
        & Plateau Steps       & $\mathcal{U}(0.00, 10.00)$  & 1.00 \\
        & Contact Friction    & $\mathcal{U}(0.10, 1.50)$   & 0.70 \\ \hline
    \multirow{4}{*}{\textbf{Stairs}} 
        & Step Size           & $\mathcal{U}(0.01, 0.50)$   & 0.10 \\
        & Maximum Height      & $\mathcal{U}(0.01, 1.50)$   & 0.01 \\
        & Number of Steps     & $\mathcal{U}(1.00, 20.00)$   & 1.00 \\
        & Contact Friction    & $\mathcal{U}(0.10, 1.50)$   & 0.70 \\ \hline
    \multirow{6}{*}{\textbf{Hilly}} 
        & Fractal Octaves     & $\mathcal{U}(0.00, 5.00)$   & 1.00 \\
        & Fractal Lacunarity  & $\mathcal{U}(1.00, 5.00)$   & 2.00 \\
        & Fractal Gain        & $\mathcal{U}(0.10, 1.00)$   & 0.45 \\
        & $z$-Scale           & $\mathcal{U}(0.01, 0.75)$   & 0.15 \\
        & Contact Friction    & $\mathcal{U}(0.10, 1.50)$   & 0.70 \\
        & Number of Samples   & $\mathcal{U}(50, 100)$      & 50   \\ \hline
\end{tabular}
\caption{Parameter ranges and nominal values for terrain configurations. Note: For stairs terrain, we define the step height as maxHeight / noSteps.}
\label{tab:terrain_parameters}
\vspace{-0.2cm}
\end{table}

By applying an initial $SR=0.1$ for terrain configuration sampling, we expose the early stage of the training regimen to less challenging terrain configurations, for example, having fewer and smaller steps for the Stairs terrain, while increasing or decreasing the maximum step height and number of stairs as a function of training performance.

\subsection{Hyperparameter choices for PPO}\label{sec:appendix:trainingDetails}

The locomotion control policy was optimized using the Proximal Policy Optimization (PPO) algorithm~\cite{schulmanProximalPolicyOptimization2017}. Both the control policy $\pi$ and the corresponding value estimator were implemented as Multi-Layer Perceptrons (MLPs) with leaky ReLU activations, comprising three hidden layers of dimensions [512, 256, 128]. The MLP estimator network consists of layers of sizes [512, 256, 64], also with leaky ReLU activation function. An MLP is the simplest neural network architecture and, compared to other memory-based networks, is more computationally efficient and has been shown to be sufficient for estimation and perceptive locomotion~\cite{jiConcurrentTrainingControl2022a, wernerArchitectureAllYou2025}. To discourage unstable behavior, an early termination penalty of $-0.2$ was introduced whenever self-collisions or non-foot body-ground contacts occurred. Additional implementation details can be found in Table~\ref{tab:method:hyperparameters}.

Training parameters and schedules were empirically adjusted following initial values presented in~\cite{rytzReferenceFreePlatform2025}. The experiments were conducted using 8 CPU cores (4 GHz) and an NVIDIA RTX 4090 GPU. Table~\ref{tab:method:hyperparameters} outlines the hyperparameters and computational settings adopted. A training horizon of 60,000 iterations required approximately 30 hours of wall-clock time for a \SI{100}{\hertz} controller network frequency.

\begin{table}[h!]
    \centering
    \small
    \setlength{\tabcolsep}{3pt}
    \begin{tabular}{|l|c||l|c|}
    \hline
    \textbf{Parameter} & \textbf{Value} & \textbf{Parameter} & \textbf{Value} \\
    \hline
        Discount factor, $\gamma$   & 0.997     & Entropy coefficient & 0 \\
        Learning rate               & adaptive  & Value coefficient & 0.5 \\
        Batch size                  & 54000     & GAE & True \\
        Mini-batch size             & 6         & GAE $\lambda$ & 0.95 \\
        Epochs                      & 4         & Steps per iteration & 120 \\
        Parallel envs, $n_{env}$    & 450       & PPO time/iteration & $\sim$1.8 s \\
    \hline
    \end{tabular}
    \caption{RL training time and hyperparameters.}
    \label{tab:method:hyperparameters}
    \vspace{-0.5cm}
\end{table}


%% file: sections/results_and_discussion.tex
\section{Results and Discussion}\label{sec:results_and_discussion}

\subsection{Performance Benchmarks}
We evaluate the generalization and performance of zero-shot \textit{morphology-specific universal} locomotion controllers across varied quadruped morphologies in simulation and on the large-scale ANYmal hardware platform. Our comparison focuses on robustness under morphology variations and perception placement, rather than peak dynamic performance. The following architectures are considered:

\begin{enumerate}
    \item \textbf{MorAL}~\cite{luoMorALLearningMorphologically2024}\textbf{:} A blind actor combined with a perceptive critic that receives privileged state-action information. The architecture is trained using a single robot reference model (Unitree A1) and adjusts the robot configuration parameter ranges according to the original implementation. This \textit{universal} controller architecture ensures that, on hardware, no requirements on the perception pipeline exist while leveraging POMDPs to process privileged simulation information during training.
    \item \textbf{$\text{MorAL}_\text{blind}$}~\cite{rytzReferenceFreePlatform2025}\textbf{:} A completely blind actor-critic architecture trained against multiple reference models, serving as our most resource-minimal baseline. 
    \item \textbf{MorAL+ (ours):} Our proposed method, which retains the blind actor for efficient policy execution utilizing a perceptive critic of MorAL, but trained with multiple reference robot models as in $\text{MorAL}_\text{blind}$, and noise-free critic state observations \textbf{H} as in \textbf{MorAL}.
    \item \textbf{PPAL (ours):} Our second proposed method, featuring a fully perceptive actor-critic architecture trained on multiple reference models, prioritizes critic information while the actor receives noisy perception scans. The chosen noise model matches a previously capable locomotion controller for heavy-duty quadrupedal robots, as the ANYmal ~\cite{rudin2022}, used in our hardware results section. It comes with manufacturer-grade perception software that has an industrial-grade state estimator and perception stack.
\end{enumerate}

All controllers were trained with identical adaptive terrain curricula, robot configuration sampling, reward functions, and PPO hyperparameters, resulting in comparable gait structures and enabling controlled architectural comparisons.

\subsection{Simulation Results}
\begin{figure}[t]
    \centering
    \vspace{-0.1cm}
    \includegraphics[width=0.9\linewidth]{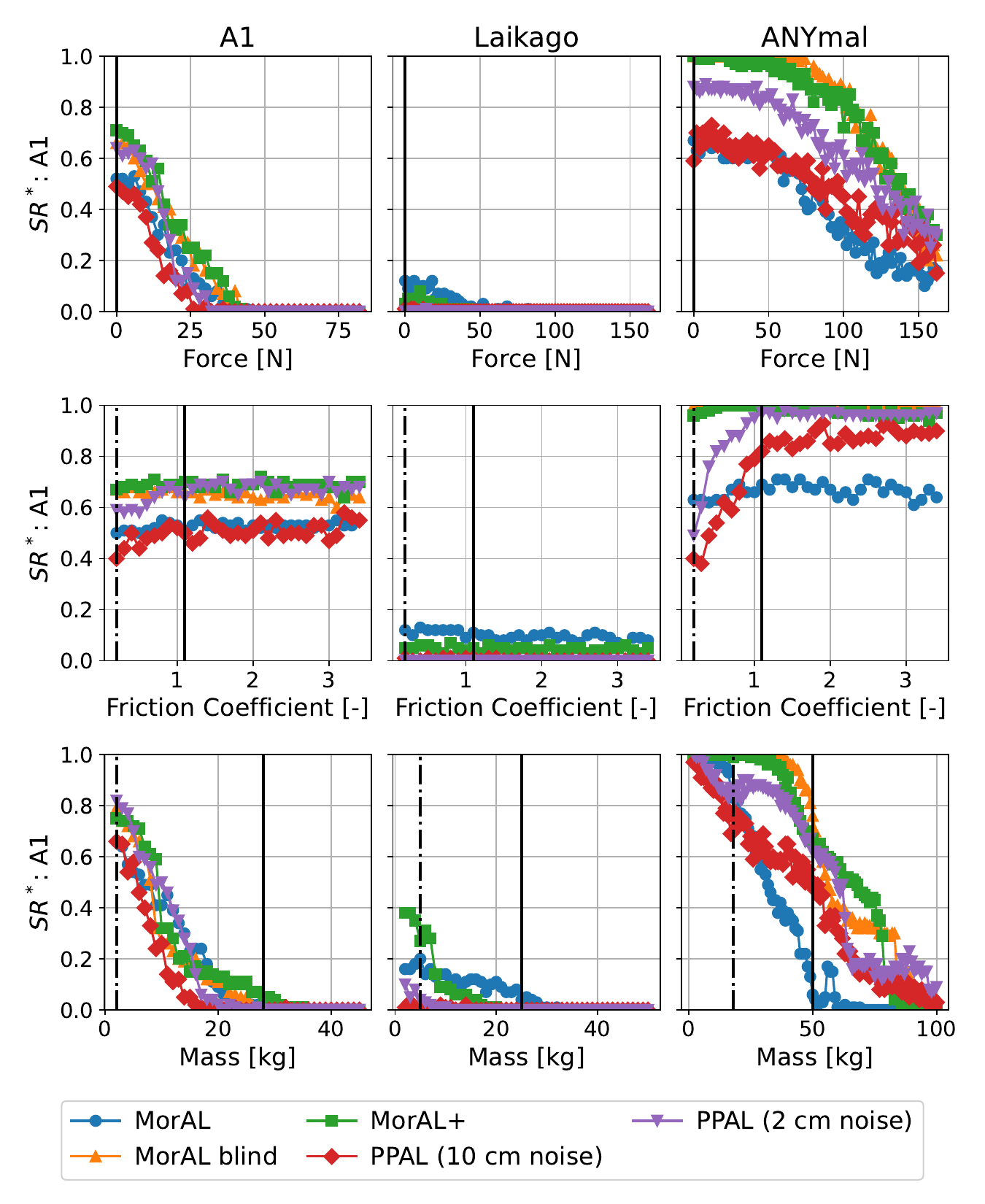}
    \caption{Robustness test of controller design with heightmap sampled at \SI{5}{\hertz} and 80\% of the training velocity command ranges. Controller policy was evaluated on quadrupeds A1, Laikago and ANYmal, previously unseen during training.}
    \label{fig:results:robotRobustness}
    \vspace{-0.3cm}
\end{figure}

We evaluate the Success Rate ($SR^*$) of our resulting estimator and actor networks in RaiSim by rolling out 100 segments of \SI{20}{\second} with random velocity commands sampled every \SI{4}{\second} of the full training commanded range~\cite{gangapurwalaLearningLowFrequencyMotion2023a}. In Fig.~\ref{fig:results:robotRobustness}, we specifically demonstrate robot configuration parameter robustness by randomizing base mass, perturbation, and contact friction effects for flat terrain locomotion. Incorporating perceptual information can degrade controller performance as demonstrated by PPAL with \SI{10}{\centi\meter} scan noise.

Consistent with~\cite{rytzReferenceFreePlatform2025}, multi-morphology training significantly improves robustness across disturbance magnitudes. Both $\mathrm{MorAL}_{\mathrm{blind}}$ and MorAL+ achieve the highest $SR^*$ over wide parameter ranges, indicating that robot configuration diversity is a primary contributor to robustness. The addition of exteroceptive input in MorAL+ does not yield a consistent performance gain over the blind variant, suggesting that the chosen formulation does not effectively leverage the high scan information. In contrast, PPAL degrades under perception noise, suggesting that directly incorporating exteroceptive input in the form of an imperfect height scan is not sufficiently denoised by the policy architecture. This indicates that more structured belief representations or memory-based inference mechanisms may be required~\cite{mikiLearningRobustPerceptive2022} to reliably denoise $\mathbf{H}^n$.

As illustrated by the Laikago results, however, none of the evaluated approaches achieve consistently high $SR^*$ across all tested perturbations. This suggests that substantial morphological shifts beyond the training distribution still degrade locomotion performance, highlighting the remaining gap in out-of-distribution morphology generalization.

\begin{figure}[h]
    \centering
    \includegraphics[width=0.85\linewidth]{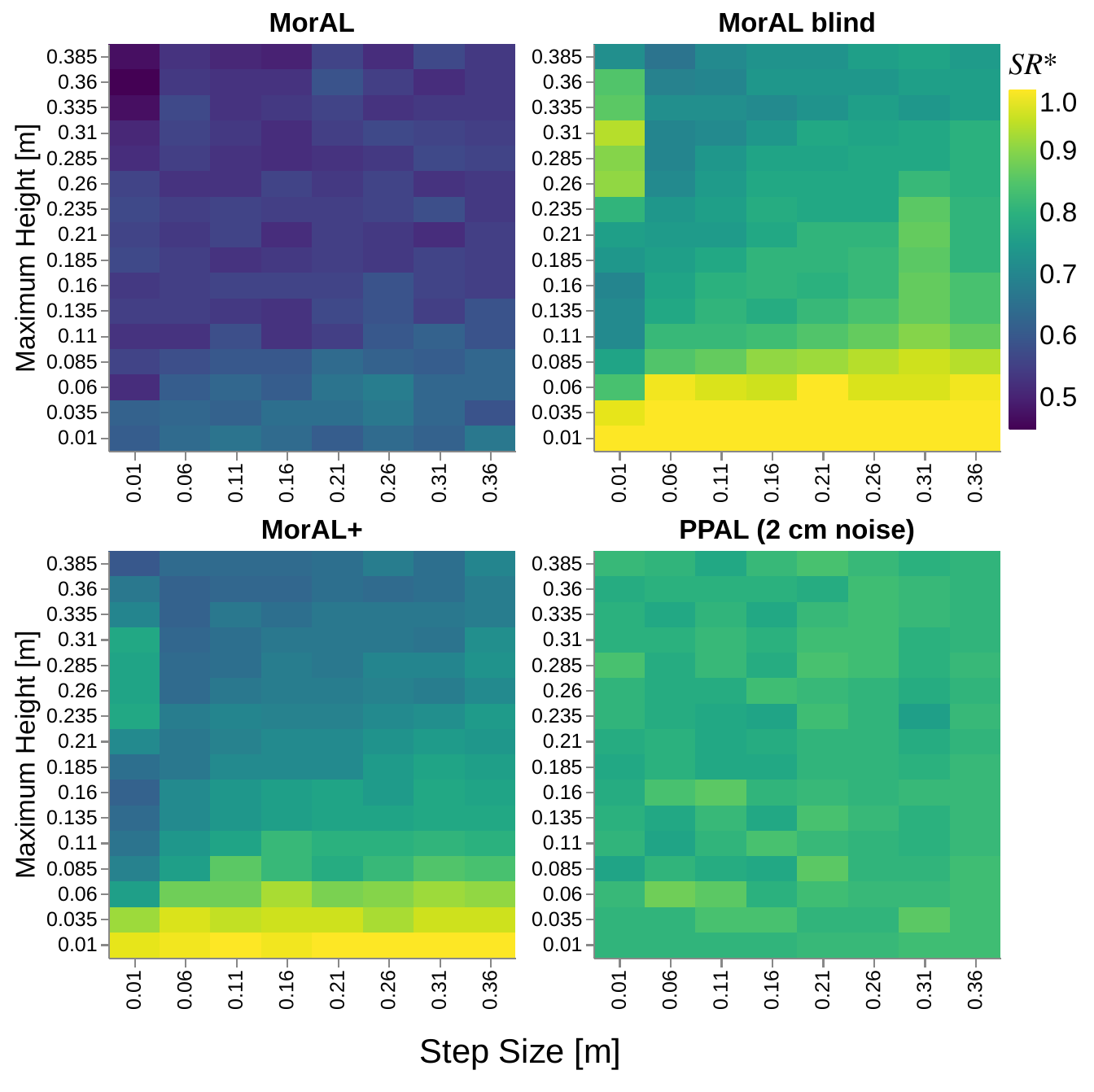}
    \vspace{-4mm}
    \caption{$SR^*$ for stair terrain and five steps over a range of maximum height and step size for ANYmal quadruped in RaiSim.}
    \label{fig:results:raisim_stair_5steps}
    \vspace{-0.4cm}
\end{figure}

We further evaluate the controllers' performance across a five-step stair terrain parameter range in Fig. ~\ref{fig:results:raisim_stair_5steps}, which presents $SR^*$ over a grid of stair step heights and lengths. The general trends observed on flat terrain transfer to stair traversal: MorAL exhibits the lowest performance due to single-morphology training, while both $\mathrm{MorAL}_{\mathrm{blind}}$ and MorAL+ achieve substantially higher success rates. Notably, $\mathrm{MorAL}_{\mathrm{blind}}$ demonstrates the most consistent performance and attains the highest success rates over a broader region of the stair parameter space. This further supports the conclusion that multi-reference morphology training is the primary driver of robustness in the proposed universal locomotion controller. In contrast, PPAL remains more sensitive to variations in stair geometry, highlighting the trade-off between incorporating perceptual inputs and maintaining robustness under imperfect or noisy exteroception.

\subsection{Hardware Results}

Tables~\ref{tab:results:pal_and_moral_combined} and~\ref{tab:results:pal_and_moral_combined_weighted} summarize hardware performance on ANYmal. MorAL+ consistently reduces velocity tracking Root Mean Square Error (RMSE) relative to $\mathrm{MorAL}_{\mathrm{blind}}$ on both flat and rough terrain, demonstrating that critic-only perception improves tracking consistency without increasing deployment complexity. PPAL achieves competitive simulation performance (see Fig.~\ref{fig:results:robotRobustness}) but exhibits increased sensitivity to perception noise, leading to less reliable hardware behavior. The original MorAL architecture could not be safely deployed due to instability (see supplementary video).


\begin{table}[h]
\vspace{-0.15cm}
\centering
\footnotesize
\setlength{\tabcolsep}{3pt} 
\renewcommand{\arraystretch}{0.95}

\begin{tabular}{l c c c c c c c}
\toprule
\textbf{Controller} & \textbf{Terrain} & 
\multicolumn{6}{c}{\textbf{RMSE}} \\
\cmidrule(lr){3-8}
 &  & $\mathbf{v}_x$ & $\mathbf{v}_y$ & $\mathbf{v}_\theta$ 
 & $\mathbf{e}_x$ & $\mathbf{e}_y$ & $\mathbf{e}_z$ \\
\midrule

    $\mathrm{MorAL}_{\mathrm{blind}}$ & Flat  & 0.0987 & 0.0955 & 0.2548 & 0.1118 & 0.1276 & 0.0702 \\
    MorAL+ & Flat & \textbf{0.0864} & \textbf{0.0909} & \textbf{0.2051} & \textbf{0.0914} & \textbf{0.1202} & \textbf{0.0579} \\
    PPAL   & Flat & 0.1605 & 0.1167 & 0.2975 & 0.1251 & 0.1213 & 0.0708 \\
    \midrule
    $\mathrm{MorAL}_{\mathrm{blind}}$ & Rough & 0.0983 & 0.0854 & 0.2734 & 0.0951 & 0.0937 & 0.0755 \\
    MorAL+ & Rough & \textbf{0.0908} & \textbf{0.0757} & \textbf{0.2401} & \textbf{0.0806} & 0.0912 & \textbf{0.0620} \\
    PPAL   & Rough & 0.1376 & 0.0762 & 0.2745 & 0.0830 & \textbf{0.0847} & 0.0690 \\
\bottomrule
\end{tabular}
    \caption{Velocity command tracking and estimation results (RMSE represented by $\mathbf{v}_x, \mathbf{v}_y, \mathbf{v}_\theta$) and base linear velocity estimate ($\mathbf{e}_x, \mathbf{e}_y, \mathbf{e}_z$) on the hardware ANYmal (\SI{5}{\minute} duration for data collection).}
    \label{tab:results:pal_and_moral_combined}
    \vspace{-0.35cm}
\end{table}

As we did not have access to an additional hardware platform to evaluate morphology transfer directly, we repeated the experiments by modifying the physical properties of the same robot. Specifically, we introduced a top-heavy payload of~\SI{6}{\kilo\gram}, thereby altering both the total mass and the center of mass of the system. The corresponding results are reported in Table~\ref{tab:results:pal_and_moral_combined_weighted}. We did not repeat the weighted experiments for PPAL, as the unweighted hardware evaluation already exhibited sagging and degraded velocity-tracking performance on flat terrain (Table~\ref{tab:results:pal_and_moral_combined}), indicating limited robustness and posing an increased risk of hardware damage under additional load.

\begin{table}[h]
\vspace{-0.15cm}
\centering
\footnotesize
\setlength{\tabcolsep}{3pt} 
\renewcommand{\arraystretch}{0.95}

\begin{tabular}{l c c c c c c c}
\toprule
\textbf{Controller} & \textbf{Terrain} & 
\multicolumn{6}{c}{\textbf{RMSE}} \\
\cmidrule(lr){3-8}
 &  & $\mathbf{v}_x$ & $\mathbf{v}_y$ & $\mathbf{v}_\theta$ 
 & $\mathbf{e}_x$ & $\mathbf{e}_y$ & $\mathbf{e}_z$ \\
\midrule

    $\mathrm{MorAL}_{\mathrm{blind}}$ & Flat  & 0.1046 & \textbf{0.0840} & 0.2488 & 0.1289 & 0.1438 & 0.0738 \\
    MorAL+ & Flat & \textbf{0.0889} & 0.0981 & \textbf{0.2470} & 0.1187 & 0.1240 & 0.0644 \\
    PPAL   & Flat & 0.1654 & 0.1191 & 0.2973 & \textbf{0.1146} & \textbf{0.1083} & \textbf{0.0598} \\
    \midrule
    $\mathrm{MorAL}_{\mathrm{blind}}$ & Rough & 0.1098 & \textbf{0.0912} & \textbf{0.2626} & 0.1300 & 0.0971 & 0.0739 \\
    MorAL+ & Rough & \textbf{0.1090} & 0.0985 & 0.2874 & \textbf{0.1070} & \textbf{0.0883} & \textbf{0.0647} \\
    PPAL   & Rough & - & - & - & - & - & - \\
\bottomrule
\end{tabular}
    \caption{Locomotion policy performance results for \SI{6}{\kilo\gram} payload.}
    \label{tab:results:pal_and_moral_combined_weighted}
    \vspace{-0.35cm}
\end{table}

All evaluated controllers achieve reliable locomotion on flat terrain. However, performance degrades on large vertical obstacles and stair environments in hardware experiments. A possible explanation is the absence of an explicit foot-clearance term in the underlying MorAL reward formulation. Without directly incentivizing swing height, the learned policy may adopt conservative swing trajectories with limited foot elevation, which appear sufficient for nominal terrains but result in insufficient clearance on pronounced vertical discontinuities.

The reference-free formulation deliberately avoids explicit stepping priors, enabling locomotion strategies to emerge during training and potentially improving generalizability. Nevertheless, the observed results indicate that the emergent behavior does not consistently provide adequate clearance for highly irregular terrain. Future work will investigate reward or architectural refinements that preserve this reference-free property while improving terrain-adaptive swing behavior.

\section{Conclusion}

In this paper, we presented and evaluated morphology-specific universal locomotion controllers for quadrupedal systems in simulation and on ANYmal hardware. Our study demonstrates that multi-morphology training substantially improves robustness across configuration and terrain variations. Injecting perception exclusively via the critic, as in MorAL+, enhances tracking consistency over blind baselines while maintaining deployment stability. In contrast, fully perceptive architectures increase sensitivity to noise.

Although stair and obstacle traversal remain challenging under the current reward design, the results highlight that perception placement and curriculum design are critical for scalable and deployable morphology-aware locomotion. Future work will investigate belief-state fusion, improved perception denoising, and reward shaping strategies to enhance terrain-aware behavior without compromising robustness across morphologies.

